# Modelling COVID-19 Pandemic Dynamics Using Transparent, Interpretable, Parsimonious and Simulatable (TIPS) Machine Learning Models: A Case Study from Systems Thinking and System Identification Perspectives

Hua-Liang Wei [1,2], Stephen A. Billings [1,2]

[1] *Department of Automatic Control and Systems Engineering, The University of Sheffield*    [2] *INSIGNEO Institute for in Silico Medicine, The University of Sheffield, Sheffield, S1 3JD*

**Abstract**

Since the outbreak of COVID-19, an astronomical number of publications on the pandemic dynamics appeared in the literature, of which many use the susceptible infected removed (SIR) and susceptible exposed infected removed (SEIR) models, or their variants, to simulate and study the spread of the coronavirus. SIR and SEIR are continuous-time models which are a class of initial value problems (IVPs) of ordinary differential equations (ODEs). Discrete-time models such as regression and machine learning have also been applied to analyze COVID-19 pandemic data (e.g. predicting infection cases), but most of these methods use simplified models involving a small number of input variables pre-selected based on a priori knowledge, or use very complicated models (e.g. deep learning), purely focusing on certain prediction purposes and paying little attention to the model interpretability. There have been relatively fewer studies focusing on the investigations of the inherent time-lagged or time-delayed relationships e.g. between the reproduction number (R number), infection cases, and deaths, analyzing the pandemic spread from a systems thinking and dynamic perspective. The present study, for the first time, proposes using systems engineering and system identification approach to build transparent, interpretable, parsimonious and simulatable (TIPS) dynamic machine learning models, establishing links between the R number, the infection cases and deaths caused by COVID-19. The TIPS models are developed based on the well-known NARMAX (Nonlinear AutoRegressive Moving Average with eXogenous inputs) model, which can help better understand the COVID-19 pandemic dynamics. A case study on the UK COVID-19 data is carried out, and new findings are detailed. The proposed method and the associated new findings are useful for better understanding the spread dynamics of the COVID-19 pandemic.

## 1. Introduction

The past 18 months have witnessed the devastating spread of the COVID-19, a disastrous global pandemic which has been and still is affecting almost every single person at each corner of the world. The attention paid to COVID-19 over the past 18 months categorically surpasses that to anything else. For example, when searching with the keyword "COVID-19" and the scope of "abstract" in the database of Web of Science, the number of published articles is 94026. With the same keyword and scope, the number of published articles in the Elsevier's abstract and citation database, Scopus, is over 112,000. With the same keyword, but only search with the scope of "in the title of the article", the number of articles given by Google Scholar is over 263,000, and if the scope is changed to "anywhere in the article", the number of publications becomes reaches over 4,340,000. Clearly, the numbers of publications on COVID-19 are categorically astronomically larger than those on any other single subject.

Presently, there are a huge number of publications on describing the spread dynamics of the pandemic, most of which employ the well-known susceptible infected removed (SIR) [1] and susceptible exposed infected removed (SEIR) models [2], or their variants [3, 4] to simulate the spread of the coronavirus. SIR and SEIR are continuous-time models which are a class of initial value problems (IVPs) of ordinary differential equations (ODEs). Regression and machine learning methods have also been applied to analyze COVID-19 pandemic data (e.g. predicting infection cases), but most of these methods use simplified models involving a small number of input variables pre-selected based on a priori knowledge, or use very complex models (e.g. deep learning) [5-7], merely focusing on the prediction purpose (e.g. positive case prediction) and paying little attention to the model interpretability. There have been relatively fewer studies focusing on the investigations of the inherent time-lagged or time-delayed relationships e.g. between the reproduction number (R number), infection cases, and deaths, analyzing the pandemic spread from a systems thinking and dynamic perspective. The present study, for the first time, proposes using systems thinking and system identification approach to build transparent, interpretable, parsimonious and simulatable (TIPS) dynamic machine learning models [8,9], establishing links between the R number, the infection cases and deaths caused by COVID-19. The TIPS models are developed based on the well-known NARMAX (Nonlinear AutoRegressive Moving Average with eXogenous inputs) method, which can help better understand the COVID-19 pandemic dynamics. A case study on the COVID-19 data of the UK is carried out, and the findings are as follows: 1) The number of daily infection cases (DIC) is closely related to the R number but lags R number from 12 to 42 days, and 2) The number of daily deaths is highly dependent on R and DCIC but lags R from 14 to 41 days and lags DIC from 13 to 27 days. These new findings make significant contributions to better understand the spread dynamics of the pandemic.

The remaining of this chapter is as follows. Section 2 briefly depicted the research problem. Section 3 provides a brief description of the method used for calculating the R number. Section 4 introduces the NARMAX methods. In Section 5, three cases studies based on the UK COVID-19 data are presented. Finally, the work is concluded in Section 6.

## 2. Problem Representation

This work aims to build transparent, interpretable, parsimonious and simulatable (TIPS) dynamic machine learning models, which are used to achieve two objectives: 1) Reveal the quantitative relationships of reproduction number (R number), infection cases and deaths, and 2) make prediction of daily infection cases (DIC) and the number of daily deaths (NDD) in advance of at least 12 days.

The objectives and procedure of the TIPS machine learning is graphically depicted in Figure 1. In this study, the TIPS model training procedure needs the historical records of the numbers of daily infection cases and daily deaths. The reproduction number R will be derived from the number of daily infection cases. The predictive models are built using the well-known NARMAX method [10]. The calculation of R number and the framework of the NARMAX method are presented in Section 3 and 4, respectively.

## 3. Reproduction Number

As mentioned in Section 2, the models to be built in this work involve three variables: The R number and the numbers of daily infection cases and daily deaths. The values of the R number are estimated using the method proposed in Zingano et al. [11]. The method is briefly presented as follows.

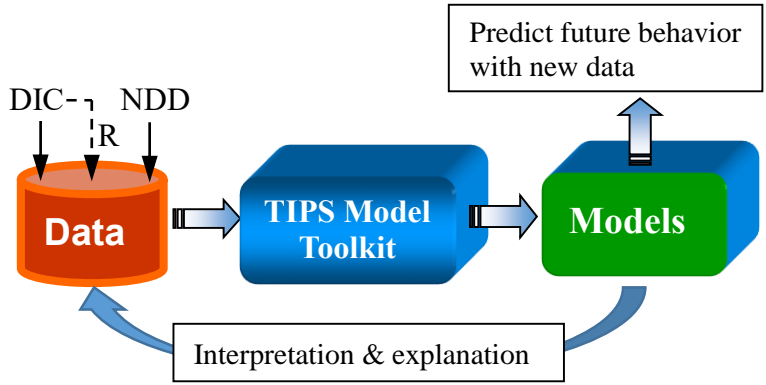

Figure 1 A diagram of the TIPS machine learning procedure

Let $S(t)$, $E(t)$, $I(t)$, $R(t)$ and $D(t)$ ($t$ can be understood in time unit of day) denote the numbers of susceptible individuals, exposed (infected but not yet infectious to transmit the disease) individuals, infectious or infected actively individuals, those recovered from the disease, and those who have died from the disease, respectively. From the mathematical theory of infectious diseases, the spread dynamics of COVID-19 can be characterized by the following SEIR model [11]:

$$\begin{aligned}
\frac{dS}{dt} &= -\beta I(t)\left(\frac{1}{N}\frac{dS(t)}{dt}\right) \\
\frac{dE}{dt} &= \beta I(t)\left(\frac{1}{N}\frac{dS(t)}{dt}\right) - \delta E(t) \\
\frac{dI}{dt} &= \delta E(t) - (r+\gamma)I(t) \\
\frac{dR}{dt} &= \gamma I(t) \\
\frac{dD}{dt} &= rI(t)
\end{aligned} \quad (1)$$

with initial values $S(t_0) = S_0$, $E(t_0) = E_0$, $I(t_0) = I_0$, $R(t_0) = R_0$, and $D(t_0) = D_0$, satisfying $S_0 + E_0 + I_0 + R_0 + D_0 = N$, where N is the full size of the susceptible population initially exposed at the initial time $t_0$. The parameters $\beta$ (average transmission rate) and $r$ (average lethality) change with time $t$ (e.g. day), and the other two parameters $\delta$ and $\gamma$ are assumed to be positive constants, which are defined as $\delta=1/T_{ALP}$ and $\gamma =1/T_{ATT}$, where $T_{AL}$ and $T_{AT}$ represent the average latent period and the average transmission time, respectively. Clinical study results show that for COVD-19, $T_{AL}$ is around 5 days on average [12] and $T_{AT}$ is around 14 days [13].

Note that the five variables, $S(t)$, $E(t)$, $I(t)$, $R(t)$ and $D(t)$, in the ordinary differential equation model (1) obey the following conservation law [11]: $S(t)+E(t)+I(t)+R(t)+D(t) = N$ (for any $t \geq t_0$).

Based on the SEIR model (1), Zingano et al. [11] developed the formula for calculating the reproduction number $R_t$ as follow:

$$RN(t) = \frac{1}{N}\frac{\beta(t)S(t)}{r(t)+\gamma} \quad (2)$$

## 4. NARMAX Methods

NARMAX methods were initially developed for solving control systems engineering modelling problems, especially complex nonlinear system identification tasks, but gradually have been successfully applied to a wide range of multidisciplinary domains including medical [14], neuroscience [15], social science [16,17], climate and weather [18], space weather [19], among others [13].

NARMAX methods employ discrete-time dynamic models [20]. Specifically, assume that a system output $y$ is potentially driven by $r$ input variables, designated by $u_1, u_2, \ldots, u_r$, the general form of the NARMAX model is:

$$y(t) = f[y(t-\tau_y), \cdots, y(t-n_y), u_1(t-\tau_u), \cdots, u_1(t-n_u), \ldots, \\ u_r(t-\tau_u), \ldots, u_r(t-n_u), e(t-1), \cdots, e(t-n_e)] + e(t) \quad (3)$$

where $y(t)$, $u(t)$ and $e(t)$ are the measured system output, input and noise sequences respectively at time instant $t$; $n_y$, $n_u$, and $n_e$ are the maximum lags for the system output, input and noise; $\tau_y$ and $\tau_u$ are a time delay between the input and output, and usually $\tau_y = \tau_u = 1$; $f[\bullet]$ is some non-linear function that needs to be estimated from measured or observed input and out data. Note that the noise $e(t)$ is unmeasurable but can be replaced by the model prediction error in system identification procedure. The noise terms are included to accommodate the effects of measurement noise, modelling errors, and/or unmeasured disturbances.

In practice, NARMAX models can be implemented using different approaches, such as recurrent neural networks [21], radial basis function (RBF) networks [22, 23], wavelet neural networks [24, 25], along with others [13]. More than often, the polynomial representation, due to its attractive interpretation properties, is employed to implement NARMAX models [26-28]. In this study, the power-form polynomial basis is considered.

NARMAX model identification usually starts from a specified dictionary or library, consisting of a sufficiently large number of candidate model elements (e.g. model terms or regressors), each of which is formed by the lagged system input and output variables, such as $y(t-1)$, $u_1(t-2)$, $u_2(t-1)$, $u_1(t-1)u_2(t-1)$. A model construction algorithm (or a combination or ensemble of a several algorithms) can then be performed on the dictionary together with a training dataset, to construct sparse or parsimonious models. For example, for a single-input single-output (SISO) system, set $n_y = 1$, $n_u = 1$, $n_e = 0$, $\tau_y = \tau_u = 1$, the candidate dictionary with the nonlinear degree $\ell = 3$ is:

$$D = \begin{cases} y(t-1), & u(t-1), & y^2(t-1), & y(t-1)u(t-1), & u^2(t-1) \\ y^3(t-1), & y^2(t-1)u(t-1), & y(t-1)u^2(t-1), & u^3(t-1) \end{cases} \quad (4)$$

The final identified model may be of the form:

$$y(t) = ay(t-1) + by(t-1)u(t-1) + cu^2(t-1) \quad (5)$$

An efficient and commonly used algorithm for NARMAX model construction is the well-knows forward regression with orthogonal least squares (FROLS) [13, 29] and its variants, e.g., minimization error minimization [30], iFROLS (iterative FROLS) [31], uFROLS (ultra-FROLS) [32]. Recently, the LASSO algorithm [33] has also been applied to system identification [34] and feature selection for classification tasks [35] but it turned out that LASSO did not outperform FROLS, because "the lasso is not a very satisfactory variable selection method in the $p \gg n$ case" where $n$ is the number of observations and $p$ is the number of predictors [36]. However, $p \gg n$ is a common case in many real applications. It was theoretically proved by Johnson et al. [37] that the "optimal" $L_1$–norm solutions are often inferior to $L_0$–norm solutions found using stepwise regression; they also compared algorithms for solving these two problems and showed that although $L_1$–norm solutions can be efficient, the "optimal" $L_1$–norm solutions are often inferior to $L_0$–norm solutions found using greedy classic stepwise regression.

The FROLS algorithm uses a simple but efficient index, called error reduction ratio (ERR) [13, 29], to measure the importance or significance of each candidate model term (element) included in the specified dictionary or library, and determine which ones should be included in the model in order of their importance, e.g., in terms of the contributions the can make to explaining the variation of the target signal (system output). The model construction procedure usually leads to transparent, interpretable, parsimonious and simple/simulatable (TIPS) models. A rigid model validation approach [38, 39] can guarantee the validity of the final identified model to sufficiently represent the input-output relationship hidden in the data.

Taking the case study, on the UK Understanding Society (UKUS) data, presented in [17] as an example, the identified models using the NARMAX methods show that, on the collective national level of the UK, the factors that appear to have significantly positive impact on happiness are: 'income (living comfortably),' 'income (doing alright),' 'income (just about getting by),' 'retired,' 'health (excellent),' 'health (every good).'; the combination of the two variables of 'retirement' and 'above 65' is also an important model term, which can be explained that that the retired people who are over 65 years old are more likely to be happy. The model also revealed that marriage could enhance the positive relationship between good health status and happiness, while smoke could enhance the negative effect of low income on happiness.

More detailed descriptions of NARMAX methods can be found in [13]. This work uses a 10 fold-cross validation scheme and the FROLS algorithm to build TIPS models. In doing so, the entire dataset is first split two parts (for training and testing, respectively), and the training data are then split two parts (around 70%:30%) which are used for model structure selection and performance validation. The standard version of the FROLS algorithm is described in detail in the Appendix.

## 5. Case Studies

The section provides case studies on the UK COVID-19 data modelling and analysis using the TIPS modelling approach based on the NARMAX methods. The two main objectives are: 1) To establish the relationships between the R number, the daily infection cases and deaths; 2) To make predictions of the daily infection cases and deaths in advance of more than 10 days.

The data were extracted from the Johns Hopkins Coronavirus Resource Center (https://coronavirus.jhu.edu/about/how-to-use-our-data). For all the case studies, a total of 529 daily data (infection cases and deaths), from 4 March 2020 to 15 August 2021, are considered for the analysis purpose here, of which the first 361 data (4 March 2020 – 28 Feb 2021) are used for model training, estimation and validation, and the remaining 168 data (1 March 2021 – 15 August 2021) are used for model performance testing.

### 5.1 The Impact of the R Number on Daily Infection Cases

To examine and investigate the impact of the daily R number values on the daily infection cases in later days, the daily R number is treated as an input and the daily infection cases as the output (response), and the settings for the NARMAX model (1) are: $n_u = 42$ (maximum time lag in input), $n_y = 0$ (without including lagged autoregressive term), $n_e = 0$ (without including noise term), $\tau_u = 1$ (time delay). The identified model is:

$$y(t) = 3.5551 \times 10^4 u(t-12) - 6.2117 \times 10^3 u(t-40) - 1.17395 \times 10^4 \quad (6)$$

where $y$ = 'daily infection cases' and $u$ = 'daily R number'. The model was simulated driven by the 513 daily data. A comparison between the model predicted values and the real data, over the 361 training data points (4 March 2020 – 28 Feb 2021) and 168 test data (1 March 2021 – 15 August 2021), is shown in Figure 2.

Model (6) indicates that the R number value of the current day may impact the daily infection cases of 12 days later, lasting until 40 days. This is also reflected in Figure 2. This important finding has not been noticed in any previous study.

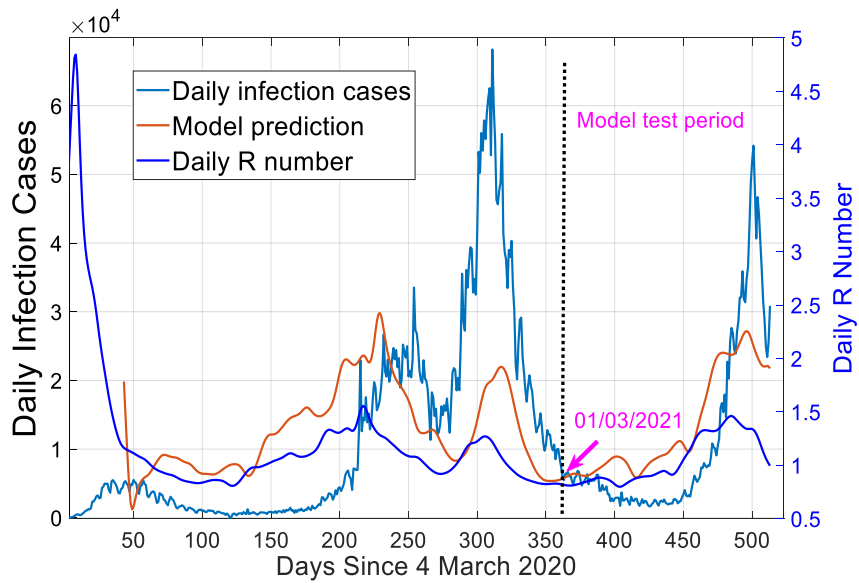

Figure 2 A comparison between the model predicted values of daily infection cases and R number from model (6) and the real data .

### 5.2 Predicting Daily Infection Cases Using R Number

Using the information given in by model (6) setting $n_u = 42$ (maximum time lag in input), $n_y = 42$ (maximum lag in output), $n_e = 0$ (without including noise term), $\tau_y = \tau_u = 12$ (time delay) and the nonlinear degree $\ell = 2$, a best NARMAX model was identified, which is shown in Table 2.

The model was simulated over the whole 513 data (4 March 2020 to 30 July 2021). The values of R-square (the coefficient of determination) on the 361 training data points (4 March 2020 – 28 Feb 2021) and 168 test data points (1 March 2021 – 15 August 2021) are 0.8991 and 0.8544, respectively. A comparison between the model predicted values and the corresponding records, over the training and test data, is shown in Figure 3.

**Table 1** The model for daily infection cases, where *y* = 'daily infection cases' and *u* = 'daily R number'.

| Index | Model Term | Parameter | ERR (100%) | P-value |
|---|---|---|---|---|
| 1 | u(t-12)×y(t-12) | 1.1674e+01 | 81.9265 | 6.4150e-11 |
| 2 | y(t-12)×y(t-18) | 8.9681e-06 | 6.4553 | 1.5963e-02 |
| 3 | u(t-13)×y(t-12) | -1.1246e+01 | 1.0121 | 6.3366e-10 |
| 4 | u(t-12)×y(t-18) | 1.0908e+00 | 1.8987 | 5.1187e-09 |
| 5 | u(t-27)×y(t-18) | -1.1120e+00 | 1.1427 | 9.9920e-15 |
| 6 | y(t-14) | 1.4192e+00 | 0.5034 | 0 |
| 7 | y(t-12)×y(t-14) | -2.6652e-05 | 0.5831 | 2.5684e-12 |
| 8 | u(t-42)×y(t-36) | -1.7229e-01 | 0.7061 | 2.1496e-09 |

Both the quantitative results (e.g. the R-square values) and graphical illustration show that the identified model show excellent prediction results. More importantly, it can be seen from Table 1 how the values of the R number and the infection cases can potential affect the pandemic spread after 12 days lasting until 42 days. For example, the combination of the two quantities of R number and infection cases of current day can potentially have a very high impact on the infection cases 12 days later, as the cross product term of the two variables has a very high ERR value, showing that it can explain nearly 90% of the variance of the daily infection cases after 12 days.

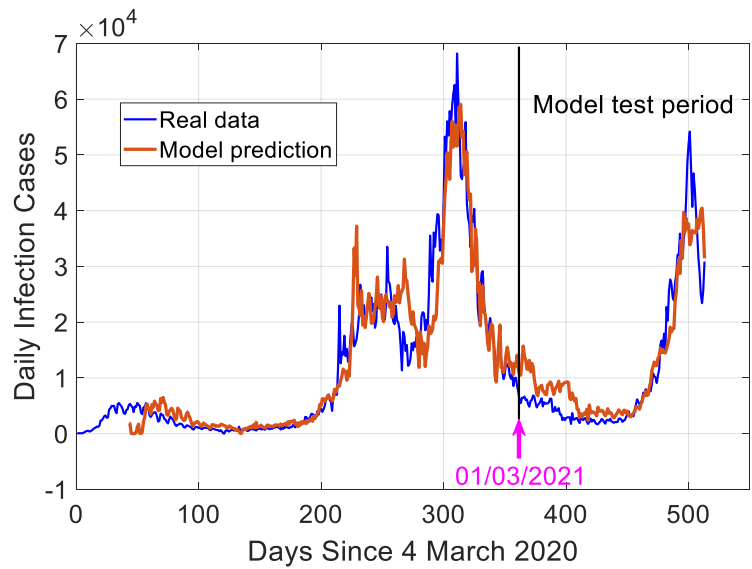

Figure 3 A comparison between 12 days ahead predictions of daily infection cases and the real data.

### 5.3 Predicting Daily Mortality Using Daily Infection Cases and R Number

In this third case, the settings as follows. The R number and daily infection cases are used as two inputs, and the number of daily deaths is considered to be the output. The other coefficients of the NARMAX (1) are chosen as: $n_u = 42$ (for both inputs), $n_y = 0$ (no autoregressive variable is included in the model), $n_e = 0$ (without including noise term), $\tau_u = 12$ (time delay) and the nonlinear degree $\ell = 2$. The identified NARMAX model is shown in Table 2. Note that the model identification algorithm did not find any nonlinear models that outperforms the linear model shown in Table 1, this suggests or implies that there is no or very weak nonlinear relation along the input and output variables; the relationship is dominated by linearity.

**Table 2** The model for daily infection cases, where *y* = 'number of daily deaths', $u_1$ = 'daily R number', and $u_2$ = 'daily infection cases'.

| Index | Model Term | Parameter | ERR (100%) | P-value |
|---|---|---|---|---|
| 1 | u2(t-13) | 1.7498e-02 | 75.9261 | 0 |
| 2 | u1(t-41) | 2.4864e+02 | 7.3788 | 0 |
| 3 | u1(t-12) | -2.3836e+02 | 4.0250 | 0 |
| 4 | u2(t-21) | 1.1513e-02 | 0.3910 | 3.6707e-04 |
| 5 | u2(t-24) | -1.3255e-02 | 0.5640 | 4.6917e-06 |
| 6 | u2(t-27) | 6.3630e-03 | 0.2339 | 1.2048e-02 |

It can be observed from Table 2 that the values of the R number (input $u_1$) can potential affect the mortality after 12 days lasting until 41 days, whereas the daily infection cases ($u_2$) can potential affect the mortality after 13 days lasting until 27 days.

From Table it can be noticed that the daily infection cases of current day (with ERR = 75.9%) potentially highly affect the number of deaths 13 days later.

The model was simulated over the entire data (4 March 2020 to 15 August 2021). A comparison between the model predicted values and the corresponding records, over the training and test data, is shown in Figure 3, where it can be seen that the model performs excellent on the training data and most part of the test data (e.g. until 21 June 2021), with the value of R-square is close to 0.8).

However, it can be clearly seen that the model fails to predict the number of daily deaths in most recent days (e.g. around 21 June 2021 and onward), although the daily infection cases is still very high as shown in Figure 3. This is reasonable and may probably be explained that more and more people have received a second vaccine which has helped significantly reduce the death rate, and this confirms the effectiveness of the UK government's vaccination policy.

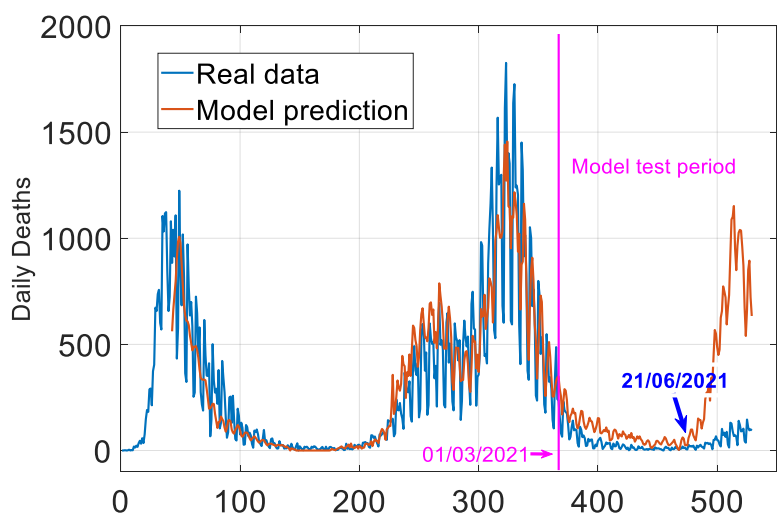

Figure 4 A comparison between 12 days ahead predictions of daily deaths and the real data.

It is worth mentioning the following two points: 1) Under the settings of $n_u = 42$, $n_y = 0$, $n_e = 0$, $\ell = 2$, the best variables that the model training algorithm identified are u2(t-13), u1(t-14), etc. This suggests that the inclusion of any other variables cannot enhance the explanation of the variation of the daily death cases using the daily infection cases and the daily value of R number, and therefore cannot help improve the prediction performance. 2) It is straightforward to detect the periodical change of the daily death cases (with a period of 7 days), the correlation between the lagged variable y(t-7) and the original signal is as high as 0.93. The main purpose of this section is to reveal the inherent dynamics that project daily infection cases and R-number to daily death cases many days later. To avoid the impact of the strong autocorrelation on the analysis of the underlying dynamics between the input and output variables, the lagged autoregressive variables, such as y(t-1), y(t-2), etc. were deliberately not considered when constructing the model in this case study.

## 6. Conclusions

The prediction of the COVID-19 pandemic is important and challenging. However, a complicated black-box that lacks interpretability (e.g. without explicitly providing information on the inherent dynamics) may become less useful or powerful for applications where there is a need to know the relationship of the inherent dynamics. The main attention of this work was paid to developing a glass-box modelling approach. The main contributions of the work can be summarized as follows:

Firstly, it proposed a TIPS-ML framework based on the NARMAX methods, and applied the proposed approach to modelling the spread dynamics based on the UK COVID-19 data. In comparison with other complicated machine learning methods, the proposed method has several highly attractive properties, such as transparency, interpretability, parsimony, and simplicity/simulatability. These properties are very important for investigating and understanding the spread dynamics of the pandemic, which may not be able to obtained by using other machine learning methods (e.g. those complicated black-box neural network models).

Secondly, some important new findings have been obtained from the identified TIPS models. For example, the R number of the current day may significantly impact the daily infection cases 12 days and last as long as 42 days; the combinational effect of R number and infection cases of current day can be potentially very high on the infection cases 12 days later; the number of daily deaths is highly dependent on R and daily infection cases (DIC) but lags R from 14 to 41 days and lags DIC from 13 to 27 days. These new findings, which have not been observed before, are useful for better understanding the spread dynamics of the pandemic.

The case studies carried out in this work focused on the UK COVID-19 data. In future, more data of different countries will be considered and analyzed using the proposed method will be applied, to investigate and compare the pandemic spread dynamic patterns, from which to acquired information that may be useful for healthcare and infectious disease studies.

## 7. Acknowledgements

This work was supported in part by the Natural Environment Research Council (NERC) under the Grant NE/V001787 and Grant NE/V002511, the Engineering and Physical Sciences Research Council (EPSRC) under Grant EP/I011056/1, the EPSRC Platform Grant EP/H00453X/1, and European Research Council under the Grant FP7-IDEAS-ERC (Grant agreement ID: 226037).

**Appendix: The Forward Regression with Orthogonal Least Squares (FROLS) Algorithm**

The TIPS models are built using a 10 fold-cross validation scheme and based on the FROLS algorithm. Taking a single-input, single-output as an example, an initial TIPS model can easily be converted into a linear-in-the-parameters form:

$$y(t) = \sum_{m=1}^{M} \theta_m \psi_m(t) + e(t) \tag{A1}$$

where $\psi_m(t) = \psi_m(\mathbf{x}(t))$ are the model regressors, $\mathbf{x}(t) = [x_1(t), x_2(t), \cdots, x_n(t)]^T$ is a vector of model 'input' variables, each element $x_i(t)$ is either one of the $n$ lagged variable such as $y(t-1)$, $y(t-2)$, …, $y(t-n_y)$, $u(t-1)$, $u(t-2)$, …, $u(t-n_u)$ ($n=n_y+n_u$), or cross-product of these lagged variables such as $y(t-12)u(t-12)$ and $y(t-12)y(t-18)$; $\theta_m$ are the model parameters, and $M$ is the total number of candidate regressors.

The initial regression model (A1) often involves a large number of candidate model terms. Experience suggests that most of the candidate model terms can be removed from the model, and that only a small number of significant model terms are needed to provide a satisfactory representation for most nonlinear dynamical systems. The FROLS algorithms (Billings, 2013) can be used to select significant model terms.

Consider the term selection problem for the linear-in-the-parameters model (A1). Let $\{(\mathbf{x}(t), y(t)) : \mathbf{x} \in \mathbf{R}^n, y \in \mathbf{R}\}_{t=1}^{N}$ be a given training data set and $\mathbf{y} = [y(1), \cdots, y(N)]^T$ be the vector of the output. Let $I = \{1, 2, \cdots, M\}$, and denote by $\Omega = \{\psi_m : k \in I\}$ the dictionary of candidate model terms in an initially chosen candidate regression model similar to (9). The dictionary $\Omega$ can be used to form a variant vector dictionary $\mathcal{D} = \{\boldsymbol{\varphi}_m : m \in I\}$, where the $k$th candidate basis vector $\boldsymbol{\varphi}_m$ is formed by the $k$th candidate model term $\psi_m \in \Omega$, in the sense that $\boldsymbol{\varphi}_m = [\psi_m(\mathbf{x}(1)), \cdots, \psi_m(\mathbf{x}(N))]^T$. The model term selection problem is equivalent to finding, from $I$, a subset of indices, $I_n = \{i_m : m = 1, 2, \cdots, n, i_m \in I\}$ where $n \leq M$, so that $\mathbf{y}$ can be approximated using a linear combination of $\boldsymbol{\alpha}_{i_1}, \boldsymbol{\alpha}_{i_2}, \cdots, \boldsymbol{\alpha}_{i_n}$.

**A.1 The forward orthogonal regression procedure**

A non-centralised squared correlation coefficient will be used to measure the dependency between two associated random vectors. The non-centralised squared correlation coefficient between two vectors $\mathbf{x}$ and $\mathbf{y}$ of size $N$ is defined as

$$C(\mathbf{x}, \mathbf{y}) = \frac{(\mathbf{x}^T \mathbf{y})^2}{\|\mathbf{x}\|^2 \|\mathbf{y}\|^2} = \frac{(\mathbf{x}^T \mathbf{y})^2}{(\mathbf{x}^T \mathbf{x})(\mathbf{y}^T \mathbf{y})} = \frac{(\sum_{i=1}^{N} x_i y_i)^2}{\sum_{i=1}^{N} x_i^2 \sum_{i=1}^{N} y_i^2} \tag{A2}$$

The squared correlation coefficient is closely related to the error reduction ratio (ERR) criterion (a very useful index in respect to the significance of model terms), defined in the standard orthogonal least squares (OLS) algorithm for model structure selection (Chen et al., 1989; Billings, 2013).

The model structure selection procedure starts from equation (A1). Let $\mathbf{r}_0 = \mathbf{y}$, and

$$\ell_1 = \arg \max_{1 \leq j \leq M} \{C(\mathbf{y}, \boldsymbol{\varphi}_j)\} \tag{A3}$$

where the function $C(\cdot,\cdot)$ is the correlation coefficient defined by (A2). The first significant basis can thus be selected as $\boldsymbol{\alpha}_1 = \boldsymbol{\varphi}_{\ell_1}$, and the first associated orthogonal basis can be chosen as $\mathbf{q}_1 = \boldsymbol{\varphi}_{\ell_1}$. The model residual, related to the first step search, is given as

$$\mathbf{r}_1 = \mathbf{r}_0 - \frac{\mathbf{y}^T \mathbf{q}_1}{\mathbf{q}_1^T \mathbf{q}_1} \mathbf{q}_1 \tag{A4}$$

In general, the $k$th significant model term can be chosen as follows. Assume that at the ($m$-1)th step, a subset $\mathcal{D}_{k-1}$, consisting of ($m$-1) significant bases, $\boldsymbol{\alpha}_1, \boldsymbol{\alpha}_2, \cdots, \boldsymbol{\alpha}_{m-1}$, has been determined, and the ($m$-1) selected bases have been transformed into a new group of orthogonal bases $\mathbf{q}_1, \mathbf{q}_2, \cdots, \mathbf{q}_{m-1}$ via some orthogonal transformation. Let

$$\mathbf{q}_j^{(m)} = \boldsymbol{\varphi}_j - \sum_{k=1}^{m-1} \frac{\boldsymbol{\varphi}_j^T \mathbf{q}_k}{\mathbf{q}_k^T \mathbf{q}_k} \mathbf{q}_k \tag{A5}$$

$$\ell_k = \arg \max_{j \neq \ell_i, 1 \leq i \leq k-1} \{C(\mathbf{y}, \mathbf{q}_j^{(k)})\} \tag{A6}$$

where $\boldsymbol{\varphi}_j \in \mathcal{D} - \mathcal{D}_{m-1}$, and $\mathbf{r}_{m-1}$ is the residual vector obtained in the ($m$-1)th step. The $m$th significant basis can then be chosen as $\boldsymbol{\alpha}_m = \boldsymbol{\varphi}_{\ell_m}$ and the $m$th associated orthogonal basis can be chosen as $\mathbf{q}_m = \mathbf{q}_{\ell_m}^{(m)}$. The residual vector $\mathbf{r}_m$ at the $m$th step is given by

$$\mathbf{r}_m = \mathbf{r}_{m-1} - \frac{\mathbf{y}^T \mathbf{q}_m}{\mathbf{q}_m^T \mathbf{q}_m} \mathbf{q}_m \tag{A7}$$

Subsequent significant bases can be selected in the same way step by step. From (A7), the vectors $\mathbf{r}_m$ and $\mathbf{q}_m$ are orthogonal, thus

$$\|\mathbf{r}_m\|^2 = \|\mathbf{r}_{m-1}\|^2 - \frac{(\mathbf{y}^T \mathbf{q}_m)^2}{\mathbf{q}_m^T \mathbf{q}_m} \tag{A8}$$

By respectively summing (A7) and (A8) for $m$ from 1 to $n$, yields

$$\mathbf{y} = \sum_{m=1}^{n} \frac{\mathbf{y}^T \mathbf{q}_m}{\mathbf{q}_m^T \mathbf{q}_m} \mathbf{q}_m + \mathbf{r}_n \tag{A9}$$

$$\|\mathbf{r}_n\|^2 = \|\mathbf{y}\|^2 - \sum_{m=1}^{n} \frac{(\mathbf{y}^T \mathbf{q}_m)^2}{\mathbf{q}_m^T \mathbf{q}_m} \tag{A10}$$

In general, the $m$th significant model term can be chosen as follows. Assume that at the ($m$-1)th step, a subset $\mathcal{D}_{m-1}$, consisting of ($m$-1) significant bases, $\boldsymbol{\alpha}_1, \boldsymbol{\alpha}_2, \cdots, \boldsymbol{\alpha}_{m-1}$, has been determined, and the ($m$-1) selected bases have been transformed into a new group of orthogonal bases $\mathbf{q}_1, \mathbf{q}_2, \cdots, \mathbf{q}_{m-1}$ via some orthogonal transformation. Let

$$\mathbf{q}_j^{(m)} = \boldsymbol{\varphi}_j - \sum_{k=1}^{m-1} \frac{\boldsymbol{\varphi}_j^T \mathbf{q}_k}{\mathbf{q}_k^T \mathbf{q}_k} \mathbf{q}_k \tag{A11}$$

$$\ell_m = \arg \max_{j \neq \ell_k, 1 \leq k \leq m-1} \{C(\mathbf{y}, \mathbf{q}_j^{(m)})\} \tag{A12}$$

where $\boldsymbol{\varphi}_j \in \mathcal{D} - \mathcal{D}_{m-1}$, and $\mathbf{r}_{m-1}$ is the residual vector obtained in the (*m*-1)th step. The *m*th significant basis can then be chosen as $\boldsymbol{\alpha}_m = \boldsymbol{\varphi}_{\ell_m}$ and the *m*th associated orthogonal basis can be chosen as $\mathbf{q}_m = \mathbf{q}_{\ell_m}^{(m)}$. The residual vector $\mathbf{r}_m$ at the *m*th step is given by

$$\mathbf{r}_m = \mathbf{r}_{m-1} - \frac{\mathbf{y}^T \mathbf{q}_m}{\mathbf{q}_m^T \mathbf{q}_m} \mathbf{q}_m \tag{A13}$$

Subsequent significant bases can be selected in the same way step by step. From (A13), the vectors $\mathbf{r}_m$ and $\mathbf{q}_m$ are orthogonal, thus

$$\|\mathbf{r}_m\|^2 = \|\mathbf{r}_{m-1}\|^2 - \frac{(\mathbf{y}^T \mathbf{q}_m)^2}{\mathbf{q}_m^T \mathbf{q}_m} \tag{A14}$$

By respectively summing (A13) and (A14) for *m* from 1 to *n*, yields

$$\mathbf{y} = \sum_{m=1}^{n} \frac{\mathbf{y}^T \mathbf{q}_m}{\mathbf{q}_m^T \mathbf{q}_m} \mathbf{q}_m + \mathbf{r}_n \tag{A15}$$

$$\|\mathbf{r}_n\|^2 = \|\mathbf{y}\|^2 - \sum_{m=1}^{n} \frac{(\mathbf{y}^T \mathbf{q}_m)^2}{\mathbf{q}_m^T \mathbf{q}_m} \tag{A16}$$

The model residual $\mathbf{r}_n$ will be used to form a criterion for model selection, and the search procedure will be terminated when the norm $\|\mathbf{r}_n\|^2$ satisfies some specified conditions. Note that the quantity $\mathrm{ERR}_m = C(\mathbf{y}, \mathbf{q}_m)$ is just equal to the *m*th error reduction ratio (Chen et al., 1989; Billings, 2013), brought by including the *m*th basis vector $\boldsymbol{\alpha}_m = \boldsymbol{\varphi}_{\ell_m}$ into the model, and that $\sum_{m=1}^{n} C(\mathbf{y}, \mathbf{q}_m)$ is the increment or total percentage that the desired output variance can be explained by $\boldsymbol{\alpha}_1, \boldsymbol{\alpha}_2, \cdots, \boldsymbol{\alpha}_n$.

Finally, a mean square error (MSE) based algorithm, e.g. Akaine's information criterion (AIC), Bayesian information criterion, generalized cross-validation (GCV) and adjustable prediction error sum of squares (APRESS) can be used to determine the model size [40].

### A.2  Parameter estimation

It is easy to verify that the relationship between the selected original bases $\boldsymbol{\alpha}_1, \cdots, \boldsymbol{\alpha}_n$, and the associated orthogonal bases $\mathbf{q}_1, \cdots, \mathbf{q}_n$, is given by

$$\mathbf{A}_n = \mathbf{Q}_n \mathbf{R}_n \tag{A17}$$

where $\mathbf{A}_n = [\boldsymbol{\alpha}_1, \cdots, \boldsymbol{\alpha}_n]$, $\mathbf{Q}_n$ is an $N \times n$ matrix with orthogonal columns $\mathbf{q}_1, \mathbf{q}_2, \cdots, \mathbf{q}_n$, and $\mathbf{R}_n$ is an $n \times n$ unit upper triangular matrix whose entries $u_{ij} (1 \leq i \leq j \leq n)$ are calculated during the orthogonalization procedure. The unknown parameter vector, denoted by $\boldsymbol{\theta}_n = [\theta_1, \cdots, \theta_n]^T$, for the model with respect to the original bases, can be calculated from the triangular equation $\mathbf{R}_n \boldsymbol{\theta}_n = \mathbf{g}_n$ with $\mathbf{g}_n = [g_1, g_2, \cdots, g_n]^T$, where $g_k = (\mathbf{y}^T \mathbf{q}_k)/(\mathbf{q}_k^T \mathbf{q}_k)$ for k=1,2, …, *n*.

The model parameters reported in Tables 1 and 2 in Section 5 are the estimated values of $[\theta_1, \cdots, \theta_n]^T$.